\documentclass[letterpaper, 10 pt, conference]{ieeeconf}  % Comment this line out if you need a4paper

\IEEEoverridecommandlockouts                              % This command is only needed if 
\usepackage{graphicx}
\graphicspath{ {./imgs/} }
\usepackage{amsmath} % assumes amsmath package installed
\usepackage{amssymb}  % assumes amsmath package installed

\usepackage{multicol}
\usepackage{multirow}
\usepackage{makecell}
\usepackage{balance}
\usepackage{setspace}
\usepackage{booktabs}
\usepackage{hyperref}

\usepackage{xcolor}
\usepackage{bm}

\usepackage{enumerate}
\usepackage{comment}
\usepackage{subcaption}
\usepackage{float}

\usepackage[noadjust]{cite}

\title{\LARGE \bf
GSFeatLoc: Visual Localization Using Feature Correspondence \\on 3D Gaussian Splatting 
}
\author{
Jongwon Lee$^{1}$ and Timothy Bretl$^{1}$
\thanks{$^{1}$Jongwon Lee and Timothy Bretl are with the Department of Aerospace Engineering, University of Illinois Urbana-Champaign, Urbana, IL 61801, USA (Email: \texttt{\{jongwon5, tbretl\}@illinois.edu}).}
}

\begin{document}

\maketitle
\thispagestyle{empty}
\pagestyle{empty}

% set spacing above/below equations
\setlength{\abovedisplayskip}{0pt} \setlength{\abovedisplayshortskip}{0pt}

%%%%%%%%%%%%%%%%%%%%%%%%%%%%%%%%%%%%%%%%%%%%%%%%%%%%%%%%%%%%%%%%%%%%%%%%%%%%%%
\begin{abstract}

% In this paper, we present a method for localizing a query image
% %--- i.e., for estimating the pose of the camera with which the image was captured ---
% with respect to a precomputed 3D Gaussian Splatting (3DGS) scene representation.
% First, the method uses 3DGS to render a synthetic RGBD image at some initial guess of image pose.
% Second, the method establishes 2D-2D correspondences between the query image and this synthetic image.
% Third, the method uses the depth map to lift the 2D-2D correspondences to 2D-3D correspondences and solves a perspective-n-point (PnP) problem to compute a pose estimate.

% given an initial guess of image pose.
% The method first establishes 2D–2D correspondences between the query image and a synthetic image rendered by 3DGS at the initial pose estimate.
% It then solves a perspective-n-point (PnP) problem using the corresponding 3D points obtained from the depth map that is also rendered by 3DGS at the initial pose estimate.

In this paper, we present a method for localizing a query image with respect to a precomputed 3D Gaussian Splatting (3DGS) scene representation.
First, the method uses 3DGS to render a synthetic RGBD image at some initial pose estimate.
Second, it establishes 2D-2D correspondences between the query image and this synthetic image.
Third, it uses the depth map to lift the 2D-2D correspondences to 2D-3D correspondences and solves a perspective-n-point (PnP) problem to produce a final pose estimate.
Results from evaluation across three existing datasets with 38 scenes and over 2,700 test images show that our method significantly reduces both inference time (by over two orders of magnitude, from more than 10 seconds to as fast as 0.1 seconds) and estimation error compared to baseline methods that use photometric loss minimization.
Results also show that our method tolerates large errors in the initial pose estimate of up to $\bm{55^{\circ}}$ in rotation and $\bm{1.1}$ units in translation (normalized by scene scale), achieving final pose errors of less than $\bm{5^{\circ}}$ in rotation and $\bm{0.05}$ units in translation on 90\% of images from the Synthetic NeRF and Mip-NeRF360 datasets and on 42\% of images from the more challenging Tanks and Temples dataset.

\end{abstract}

%%%%%%%%%%%%%%%%%%%%%%%%%%%%%%%%%%%%%%%%%%%%%%%%%%%%%%%%%%%%%%%%%%%%%%%%%%%%%%
\section{Introduction}
\label{section:Introduction}

Visual localization is the process of determining the pose (position and orientation) of a query image with respect to a previously reconstructed scene (i.e., a map).
%---has been an active area of research.
It is a key component of both low-level perception modules, such as relocalization in visual simultaneous localization and mapping (vSLAM) when tracking is lost, and high-level systems for autonomous navigation, augmented reality, robotic manipulation, and human–robot interaction~\cite{miao2024survey-vloc}.
% Among various scene representations such as point clouds, meshes, and neural radiance fields (NeRF)~\cite{mildenhall2021nerf}, 3D Gaussian Splatting (3DGS)~\cite{kerbl2023gs} stands out by offering high-quality novel view synthesis---which point clouds and meshes lack---with significantly faster training and rendering compared to NeRF, making it particularly appealing for robotics applications~\cite{zhu2024survey-3dgs-robotics}.

In this paper, we restrict our attention to visual localization with respect to a 3D Gaussian Splatting (3DGS) scene representation, in particular~\cite{kerbl2023gs}. This scene representation offers high-quality novel view synthesis---which alternatives like point clouds and meshes lack---with significantly faster training and rendering than neural radiance fields (NeRF)~\cite{mildenhall2021nerf}, making it appealing for robotics applications~\cite{zhu2024survey-3dgs-robotics}.

%In this paper, we assume that a rough initial pose estimate is available (e.g., from image retrieval, other sensory data, or auxiliary methods), where by ``rough'' we mean any pose estimate that 

% It is reasonable to assume that a rough initial pose estimate, such that the image taken at that pose would have overlap with the query image, is available in visual localization---for instance, from image retrieval, other sensory data, or auxiliary methods.
% Under this assumption, visual localization becomes the task of accurately and efficiently estimating the camera pose given an initial pose.
% In what follows, we assume such a setting.	
% Our work focuses on visual localization using 3DGS as the scene representation, with the availability of a rough initial pose estimate.

Existing 3DGS-based visual localization approaches generally estimate the camera pose by defining and minimizing the photometric loss between a rendered image at the estimated pose and the query image~\cite{sun2023icomma,jiang2024gsreloc,botashev2024gsloc,jun2024gsloc}.
However, these photometric loss minimization approaches---which rely on gradient-based optimization by iteratively rendering and comparing images---have slow inference times that are insufficient for real-time applications (e.g., significantly slower than 10 frames per second).
For instance, iComMa~\cite{sun2023icomma} requires over one second per image on an NVIDIA RTX A6000 GPU---outperforming its NeRF-based counterpart, iNeRF~\cite{yen2021inerf}, which takes over ten seconds---yet still falls short of real-time performance.

% An alternative is an end-to-end approach, which directly estimates camera pose from a query image using a learned model, such as 6DGS~\cite{matteo2024-6dgs}. Although 6DGS reports real-time inference, achieving approximately less than one-tenth of a second per image on a single NVIDIA RTX 3090 GPU, and outperforms existing NeRF-based baselines~\cite{yen2021inerf,lin2023pnerf}, such end-to-end approaches require scene-specific training, limiting their application.% across different scenes.

\begin{figure}[!tbp]
    \centering
    \captionsetup[subfloat]{labelformat=parens}
    % Column titles
    % \makebox[0.48\linewidth]{\centering Query}  \hfill
    % \makebox[0.48\linewidth]{\centering Initial Guess}\vspace{-0.7em}
    % First row
    \subfloat[Query and Initial Pose]{%
        \includegraphics[width=0.98\linewidth]{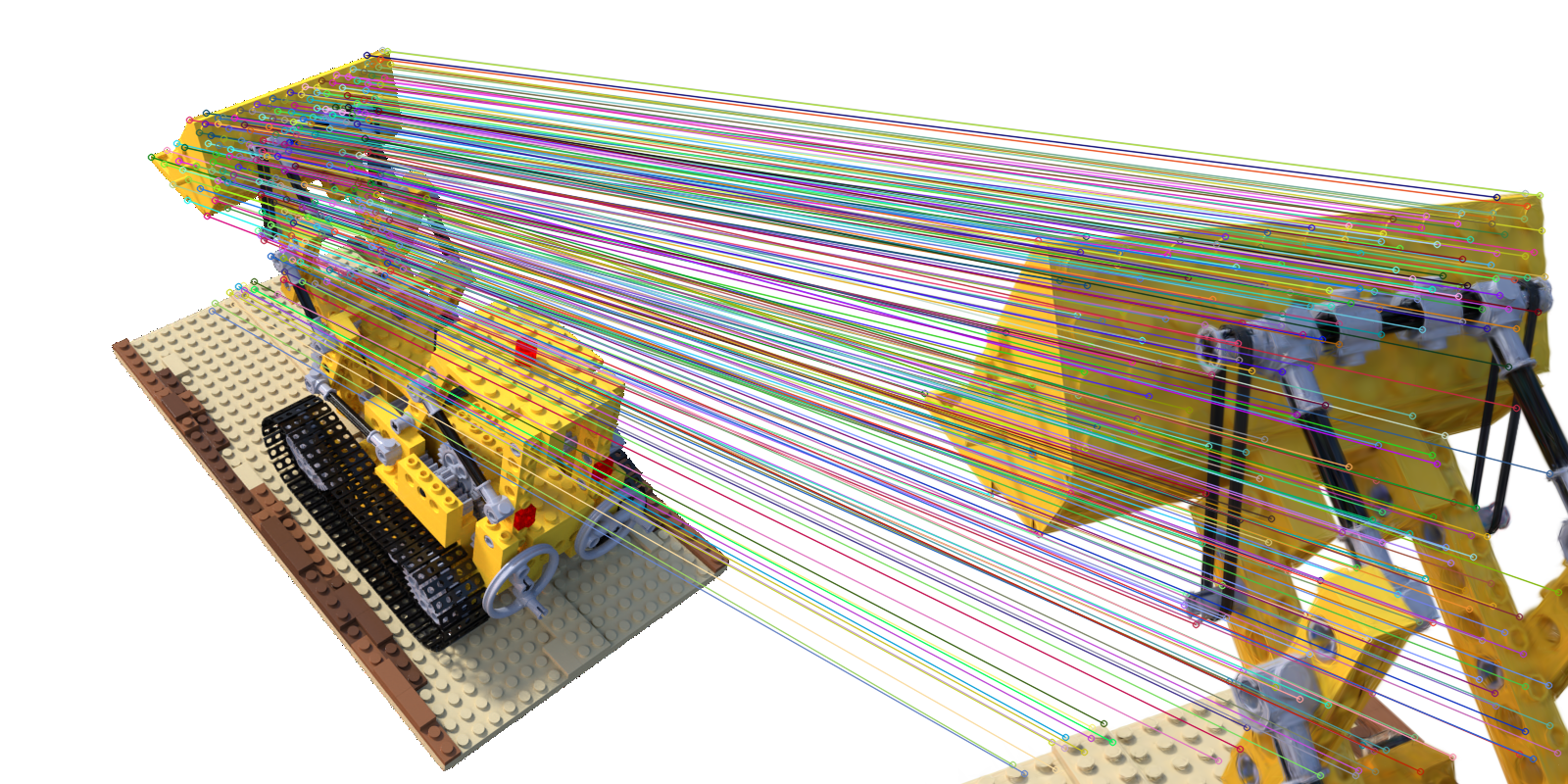}
    }\\ \vspace{0.5em}
    % \makebox[0.48\linewidth]{\centering Rendered Depth} \hfill
    % \makebox[0.48\linewidth]{\centering Estimate}\vspace{-0.7em}
    \subfloat[Rendered Dpth]{%
        \includegraphics[width=0.48\linewidth]{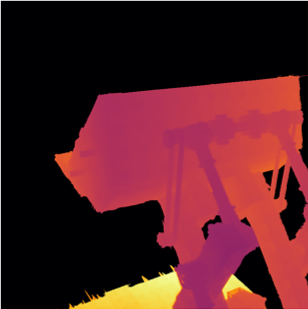}
    }\hfill
    \subfloat[Estimate]{%
        \includegraphics[width=0.48\linewidth]{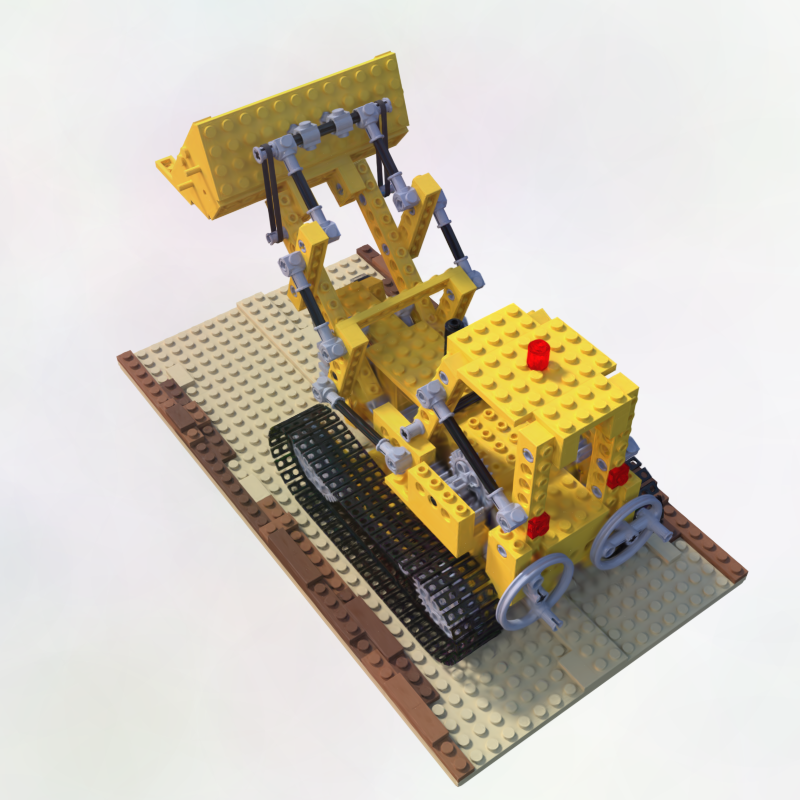}
    }% \vspace{-0.3em}
    \caption{Our method estimates the pose of a query image on a 3D Gaussian Splatting (3DGS) scene by establishing feature correspondences between the query image and an image rendered at a rough initial pose (a). The matched points are then lifted to 3D using the depth map rendered from 3DGS (b), and the final pose is estimated by solving a PnP problem. The image rendered at the estimated pose (c) is also shown.}
    \label{fig:thumbnail}
\end{figure}

To address these limitations, we propose a method of visual localization with respect to a 3DGS scene representation that needs to render only one synthetic image and that uses feature correspondence rather than photometric loss minimization (see Fig.~\ref{fig:thumbnail} for a high-level overview and Fig.~\ref{fig:diagram} for a detailed workflow).
First, our method uses 3DGS to render a synthetic RGBD image at some initial pose estimate.
Second, it establishes 2D-2D correspondences between the query image and this synthetic image.
Third, it uses the depth map to lift the 2D-2D correspondences to 2D-3D correspondences and solves a perspective-n-point (PnP) problem to produce a final pose estimate.
Conceptually, our method is a feature-based alternative to iComMa~\cite{sun2023icomma}, which uses photometric loss minimization with respect to a 3DGS scene representation, in the same way that the method of Chen \textit{et al.}~\cite{chen2024nerffeaturmatching} is a feature-based alternative to iNeRF~\cite{yen2021inerf} or PiNeRF~\cite{lin2023pnerf}, which use photometric loss minimization with respect to a NeRF scene representation. This relationship is summarized by Table~\ref{table:taxonomy}.

Although we assume that an initial pose estimate is available, the error in this initial pose estimate can be large.
%In particular, results show that our method tolerates 
% In particular, we have found that 
% ---within a range around the ground-truth such that the corresponding view overlaps with the query image---
In particular, the results we present in Section~\ref{section:Experiments} show that our method tolerates initial errors of up to $55^{\circ}$ in rotation and $1.1$ units in translation (normalized by scene scale), which is the maximum error that would still allow some overlap between the query image and the synthetic image that is rendered at the initial pose estimate.

% To address these limitations, we propose a method by establishing 2D–2D correspondences between the query image and a rendered image at an initial pose, and solving a perspective-n-point (PnP) problem using the corresponding 3D points obtained from the depth map rendered from the 3DGS, as shown in Fig.~\ref{fig:thumbnail} for a high-level overview and in Fig.~\ref{fig:diagram} for a detailed workflow.
% Conceptually, our method serves as a method based on feature correspondence alternative to iComMa~\cite{sun2023icomma}, a 3DGS-based localization approach based on minimizing photometric loss---just as the method by Chen \textit{et al.}~\cite{chen2024nerffeaturmatching} proposes a method based on feature correspondence alternative to the method based on photometric loss minimization in NeRF-based localization~\cite{yen2021inerf,lin2023pnerf}. 
% This relationship is also illustrated in Table~\ref{table:taxonomy}.
% % We note that our method is largely inspired by the approach proposed in Chen et al.~\cite{chen2024nerffeaturmatching}, which adopts feature correspondences within NeRF-based localization.
% % We show that our method consistently outperforms photometric loss–based approaches---both NeRF~\cite{lin2023pnerf} and 3DGS~\cite{sun2023icomma}---in accuracy, while achieving over two orders of magnitude gain in inference time (from tens of seconds to about 0.1 s). This is validated under various initializations (random or provided by an end-to-end method~\cite{matteo2024-6dgs}) across three datasets comprising 38 scenes and over 2,700 test images.

Our paper is structured as follows. Section~\ref{section:Related Works} reviews existing literature on visual localization methods, particularly those based on NeRF or 3DGS as scene representations. Section~\ref{section:Methodology} provides details on the proposed method based on feature correspondence. Section~\ref{section:Experiments} validates our method, in comparison with open-source baseline methods that use photometric loss minimization on NeRF or 3DGS.
We conduct experiments under two different scenarios: when the initial pose is randomly sampled (Section~\ref{subsection:Comparison under randomized initial poses} and Section~\ref{subsection:Comparison with a method based on feature correspondence using NeRF}) and when it is provided by 6DGS~\cite{matteo2024-6dgs} (Section~\ref{subsection:Comparison under initial poses provided by 6DGS}). We also analyze the sensitivity of the method with respect to error in the initial pose estimate
% the deviation of the initial pose from the ground truth
(Section~\ref{subsection:Sensitivity to initial pose from ground-truth}), and assess the effect of the choice of feature extractor and matcher (Section~\ref{subsection:Choice of feature point extractor and matcher}). 
Finally, Section~\ref{section:Conclusion} concludes the paper and discusses several directions for future work.

% The code used in this paper is available online.~\footnote{https://github.com/jongwonjlee/gsfeatloc/}
% https://anonymous.4open.science/r/gsfeatloc-F428/
% The key contributions of our work are as follows:
% \begin{itemize}
%     \item We present a method for visual localization on the 3DGS scene representation given a rough initial pose, by first establishing feature correspondences between the query image and the image rendered at the initial estimate. By lifting the 2D features to 3D coordinates using a depth map rendered by 3DGS, the final pose is estimated using the PnP algorithm.
%     \item We show that the proposed approach consistently outperforms photometric loss–based methods (both NeRF- and 3DGS-based) in terms of accuracy, and achieves significant improvements in inference time—by up to two orders of magnitude (from tens of seconds to about one-tenth of a second). This is validated under various initial poses, across three datasets comprising 38 scenes and over 2,700 test images in total.    
%     \item We open-source our implementation to facilitate future research.
% \end{itemize}

%%%%%%%%%%%%%%%%%%%%%%%%%%%%%%%%%%%%%%%%%%%%%%%%%%%%%%%%%%%%%%%%%%%%%%%%%%%%%%
\section{Related Works}
\label{section:Related Works}

The emergence of NeRF~\cite{mildenhall2021nerf} and 3DGS~\cite{kerbl2023gs}, known for their realistic scene reconstruction and novel view synthesis capabilities, has introduced new directions in visual localization.  
% Broadly, related methods can be grouped into two categories.  
% The first incorporates NeRF or 3DGS into the localization pipeline without requiring prior pose estimates, through fully end-to-end learning-based approaches (e.g., 6DGS~\cite{matteo2024-6dgs} on 3DGS as a scene representation) or by integrating these representations into specific submodules---for example, leveraging a local NeRF or 3DGS map to refine rough estimates obtained via image retrieval~\cite{moreau2023crossfire,chen2024nefes,liu2024hr-apr}. 
% This category primarily addresses global pose estimation.  
% The second assumes a rough initial pose---within a range around the ground-truth such that the image that would have taken at that pose overlaps with the query image---typically obtained from image retrieval or auxiliary sensors, and seeks to refine it within a NeRF~\cite{yen2021inerf,lin2023pnerf,chen2024nerffeaturmatching} or 3DGS~\cite{sun2023icomma,jun2024gsloc,botashev2024gsloc,jiang2024gsreloc} representation.
% This category, therefore, focuses more on local pose estimation.
% In this paper, our focus is on the latter, which explicitly performs localization given a rough initial pose.
While some recent approaches incorporate NeRF or 3DGS into localization pipelines without requiring any initial pose estimates (e.g., through fully end-to-end learning-based methods~\cite{matteo2024-6dgs}), a more common setting assumes a rough initial pose---within a range around the ground-truth such that the corresponding view overlaps with the query image---typically obtained from image retrieval or auxiliary sensors, and seeks to refine it on a NeRF~\cite{yen2021inerf,lin2023pnerf,chen2024nerffeaturmatching} or 3DGS~\cite{sun2023icomma,jun2024gsloc,botashev2024gsloc,jiang2024gsreloc} scene representation.
Our focus is on this line of work, which explicitly performs localization given a rough initial pose.

With this scope in mind, we begin by reviewing advances in visual localization with a rough initial pose using NeRF, as the advent of NeRF preceded that of 3DGS. 
NeRF facilitates approach using photometric loss minimization thanks to its ability to render novel views that were not used during reconstruction.
In this approach, the camera pose is estimated by minimizing the discrepancy between the query image and an image rendered at the estimated pose, as proposed by iNeRF~\cite{yen2021inerf} and piNeRF~\cite{lin2023pnerf}. 
The former compares the query image with an image rendered from a single initial pose, while the latter, building upon iNeRF, renders multiple images from poses sampled around the initial pose in parallel, thereby better avoiding local minima.
% This scheme is presented in methods such as iNeRF~\cite{yen2021inerf} and piNeRF~\cite{lin2023pnerf}, which iteratively optimize the pose through gradient descent by minimizing the photometric loss.
% INeRF~\cite{yen2021inerf} estimates the camera pose by minimizing photometric loss between a query image and an image rendered from a single initial pose using a NeRF.
% Building on this, piNeRF~\cite{lin2023pnerf} extends this idea by sampling multiple poses around the initial pose and optimizing them in parallel, thereby better avoiding local minima.
Meanwhile, methods that do not incorporate photometric loss fall into a separate category, namely those based on feature correspondence.
Chen \textit{et al.}~\cite{chen2024nerffeaturmatching} estimate camera pose by matching 2D feature points between a query image and an image rendered from an initial pose using a NeRF, lifting them into 3D using rendered depth, filtering out unreliable 3D points through a consistency check, and solving for pose via PnP. 
% The strategy evaluates the consistency of 3D points by rendering multiple nearby views from the initial pose and checking whether the projected positions of the points remain stable across these views.
As a post-processing step, they further refine the pose by applying photometric loss minimization (e.g., iNeRF~\cite{yen2021inerf} or piNeRF~\cite{lin2023pnerf}) with a reduced number of iterations.

Compared to NeRF, 3DGS is a more recent scene representation and has just begun to be explored for localization. 
Most existing methods rely on photometric loss~\cite{sun2023icomma,jiang2024gsreloc,botashev2024gsloc,jun2024gsloc}. 
A representative example in this category is iComMa~\cite{sun2023icomma}, which estimates camera pose by minimizing both photometric and feature-matching losses between a query image and an image rendered from an initial pose using a pre-constructed 3DGS, where the feature-matching loss is defined as the Euclidean distance between corresponding feature points on the two images.
The two losses are equally weighted in early iterations, with only photometric loss used in later iterations.
However, such a method based on photometric loss minimization requires iterative rendering and comparison---often over hundreds of steps---resulting in high computational cost and limited real-time applicability (e.g., reported inference times exceeding one second per image~\cite{sun2023icomma}).

In this work, we propose a visual localization method using 3DGS, under the assumption that a rough initial pose estimate is provided.
Our approach is analogous to that of Chen \textit{et al.}~\cite{chen2024nerffeaturmatching}, which applies classical feature correspondence techniques to a NeRF-based scene representation.
Specifically, our method uses a subset of their pipeline: unlike Chen \textit{et al.}, we do not perform the consistency-based filtering of 3D points or apply photometric loss minimization for post-pose refinement.
By using feature correspondence instead of photometric loss minimization, our method offers a significantly faster alternative for 3DGS-based localization, while not compromising estimation accuracy and even allowing a wider range of initial pose estimates.
Table~\ref{table:taxonomy} summarizes representative visual localization approaches that assume a rough initial pose and operate on NeRF or 3DGS, as discussed in this section.

\begin{table}[!htbp]
    \centering
    \caption{\textsc{Representative visual localization methods with an initial pose estimate, categorized by scene representation and pose estimation approach}}
    \begin{tabular}{c|c|c}
        \toprule
         & \multicolumn{2}{c}{\textbf{Scene Representation}} \\ \cmidrule(lr){2-3}
        \textbf{Approach} & \textbf{NeRF} & \textbf{3DGS} \\ \midrule
        \makecell{Photometric Loss \\ Minimization} & \makecell{\href{https://github.com/yenchenlin/iNeRF-public}{iNeRF}$^\dagger$~\cite{yen2021inerf}, \\ 
        \href{https://pnerfp.github.io/}{piNeRF}$^\dagger$~\cite{lin2023pnerf}}
        & \href{https://github.com/YuanSun-XJTU/iComMa}{iComMa}$^\dagger$~\cite{sun2023icomma} \\ \midrule
        \makecell{Feature \\ Correspondence} & Chen \textit{et al.}~\cite{chen2024nerffeaturmatching} & \textbf{{Ours}} \\
        % End-to-end & - & \href{https://mbortolon97.github.io/6dgs/}{6DGS}$^\dagger$~\cite{matteo2024-6dgs} \\
        \bottomrule
    \end{tabular}
    \captionsetup{width=.95\textwidth}
    \caption*{$\dagger$: Source code available}
    \label{table:taxonomy}
\end{table}

%%%%%%%%%%%%%%%%%%%%%%%%%%%%%%%%%%%%%%%%%%%%%%%%%%%%%%%%%%%%%%%%%%%%%%%%%%%%%%

\section{Methodology}
\label{section:Methodology}

\begin{figure*}[!htbp]
    \centering
    \vspace{1.5mm}
    \includegraphics[width=1.0\textwidth, keepaspectratio]{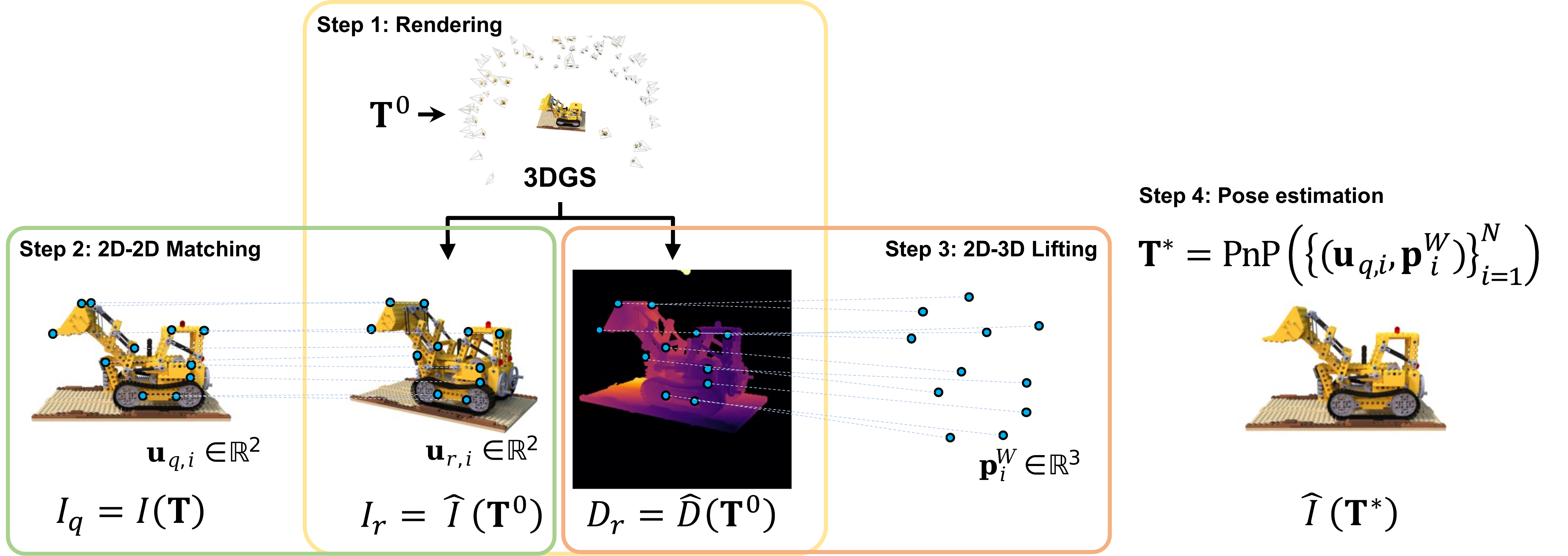}
    \caption{Overview of the proposed pipeline for visual localization using 3DGS as the scene representation, given a query image \( I_q \), an initial pose estimate \( \bm{\mathrm{T}}^{0} \in \mathrm{SE}(3)\), and the 3DGS scene representation.}
    \label{fig:diagram}
\end{figure*}

Our method estimates the camera pose of a query image by first establishing 2D–2D correspondences with a rendered image at an initial pose, and then solving a PnP problem using the corresponding 3D points obtained from the rendered depth map of the 3DGS representation (see Fig.~\ref{fig:diagram}).
% Conceptually, our method serves as a feature-based counterpart to iComMa~\cite{sun2023icomma} and GSLoc~\cite{botashev2024gsloc}, which are 3DGS-based visual localization methods that rely on photometric loss minimization.
% This is analogous to how Chen \textit{et al.}~\cite{chen2024nerffeaturmatching} serve as feature correspondence-based alternatives to NeRF-based approaches such as iNeRF~\cite{yen2021inerf} and piNeRF~\cite{lin2023pnerf}, which also minimize photometric loss.
Our method follows the approach introduced by Chen \textit{et al.}~\cite{chen2024nerffeaturmatching}, which is based on feature correspondence with NeRF as the scene representation.
Notably, our method is a simplified version of theirs, omitting both the consistency-based filtering of 3D points and the photometric loss minimization used for post-pose refinement.

\subsection{Problem statement}

Suppose we are given a query image \(I_q\),  an initial estimate of its camera pose in the world coordinate frame \(\bm{\mathrm{T}}^{W}_{C, \mathrm{init}} \in \mathrm{SE}(3)\), and a 3DGS representation of the scene. The goal of visual localization is to estimate the camera pose \(\bm{\mathrm{T}}^{W}_{C} \in \mathrm{SE}(3)\) from which the query image was captured, relative to the world coordinate frame in which the 3DGS scene is represented.
For simplicity, we omit the superscript and subscript in the coordinate notation and denote \(\bm{\mathrm{T}}^{W}_{C, \mathrm{init}}\) and \(\bm{\mathrm{T}}^{W}_{C}\) as \(\bm{\mathrm{T}}^0\) and \(\bm{\mathrm{T}}\), respectively.

\subsection{Step 1: Rendering the image and the depth map}

The first step starts by rendering the image \(I_r := \widehat{I}(\bm{\mathrm{T}}^0)\) and the depth map \(D_r := \widehat{D}(\bm{\mathrm{T}}^0)\) at the initial pose \(\bm{\mathrm{T}}^0\), respectively, where \(\widehat{I}(\cdot)\) and \(\widehat{D}(\cdot)\) denotes the rendering of the RGB and the depth image from the 3DGS.

\subsection{Step 2: Establishing 2D-2D feature correspondences}

The next step is to extract feature points from both the query image \(I_q\) and the rendered image \(I_r\), and to establish correspondences between them. As a result, a set of 2D correspondences is obtained as
\[\{ (\bm{\mathrm{u}}_{q,i}, \bm{\mathrm{u}}_{r,i})\}^{N}_{i=1},\] where \(N\) denotes the number of matched feature pairs between \(I_q\) and \(I_r\), and \(\bm{\mathrm{u}}_{q,i},~\bm{\mathrm{u}}_{r,i} \in \mathbb{R}^2\)  represent the corresponding pixel coordinates in the query and rendered images, respectively.

\subsection{Step 3: Establishing 2D–3D feature correspondences}

The next step is to lift the 2D feature points \( \bm{\mathrm{u}}_{r,i} \), where \( i \in \{1, \dotsc, N\} \), on the rendered image \( I_r \) into 3D space. This process is conducted by projecting each 2D point into the camera coordinate frame using the rendered depth map \( D_r \) and the camera intrinsic matrix \( K \), as follows:
\[
\bm{\mathrm{p}}^{C}_{i} = K^{-1} \bar{\bm{\mathrm{u}}}_{r,i} \, d,
\]
where \( d \) denotes the depth at pixel location \( \bm{\mathrm{u}}_{r,i} \), obtained from the depth map \( D_r \), and \( \bar{\bm{\mathrm{u}}}_{r,i} \) is the homogeneous representation of \( \bm{\mathrm{u}}_{r,i} \).

Each \( \bm{\mathrm{p}}^{C}_{i} \) is transformed to the world coordinate frame as
\[
\bar{\bm{\mathrm{p}}}^{W}_{i} = \bm{\mathrm{T}}^{0} \bar{\bm{\mathrm{p}}}^{C}_{i},
\]
where \( \bm{\mathrm{T}}^{0} \in \mathrm{SE}(3) \) is the initial pose estimate from which \( I_r \) and \( D_r \) are rendered, and \( \bar{\bm{\mathrm{p}}}^{W}_{i} \), \( \bar{\bm{\mathrm{p}}}^{C}_{i} \) denote the homogeneous representations of \( \bm{\mathrm{p}}^{W}_{i} \) and \( \bm{\mathrm{p}}^{C}_{i} \), respectively.

The resulting 3D point \( \bar{\bm{\mathrm{p}}}^{W}_{i} \) is thus associated with both the corresponding query feature point \( \bm{\mathrm{u}}_{q,i} \) and the rendered feature point \( \bm{\mathrm{u}}_{r,i} \), from which it was lifted.

\subsection{Step 4: Pose estimation}

Once the 2D–3D correspondences---i.e., the set of pairs 
\[\{ (\bm{\mathrm{u}}_{q,i}, \bm{\mathrm{p}}^{W}_{i})\}_{i=1}^{N}\]
% between the feature points on the query image \( \bm{\mathrm{u}}_{q,i} \) and their associated 3D coordinates in the world frame \( \bm{\mathrm{p}}^{W}_{i} \) 
are established, the camera pose $\bm{\mathrm{T}}$ from which the query image was captured can be estimated using the PnP algorithm.
We choose a PnP solver that minimizes the sum of squared reprojection errors using the Levenberg–Marquardt optimization~\cite{madsen2004lm}. A minimum of four 2D–3D correspondences is required to compute a valid solution. To increase robustness to outliers, we apply RANSAC and use only the inlier correspondences for the final pose estimation.
This process yields the final estimate of the camera pose, denoted as \( \bm{\mathbf{T}}^{\ast} \), for the query image.

%%%%%%%%%%%%%%%%%%%%%%%%%%%%%%%%%%%%%%%%%%%%%%%%%%%%%%%%%%%%%%%%%%%%%%%%%%%%%%

\section{Experiments}
\label{section:Experiments}

We evaluated our method against several baselines for visual localization using either NeRF or 3DGS scene representations, as summarized in Table~\ref{table:taxonomy}.
We selected piNeRF~\cite{lin2023pnerf} and iComMa~\cite{sun2023icomma} as representative open-source methods based on photometric loss minimization using NeRF and 3DGS, respectively.
We also compared our method to that of Chen \textit{et al.}~\cite{chen2024nerffeaturmatching}, which uses feature correspondence like we do but which uses NeRF instead of 3DGS as the scene representation.
Since an open-source implementation of Chen \textit{et al.}~\cite{chen2024nerffeaturmatching} was not available, the nature of our comparison in that case was different, and so is discussed separately in Section~\ref{subsection:Comparison with a method based on feature correspondence using NeRF}.
%is based on feature correspondence using NeRF---which our method follows but is not open-sourced---in Section~\ref{subsection:Comparison with a method based on feature correspondence using NeRF}.
All three methods address the same problem setting: estimating the camera pose from a single query image given a pre-constructed scene and an initial pose estimate, making them suitable for comparison.
We also did experiments to assess the value of using 6DGS~\cite{matteo2024-6dgs}, an end-to-end method of visual localization with respect to 3DGS that does not require initial pose estimates, as a source of such estimates for other methods in Section~\ref{subsection:Comparison under initial poses provided by 6DGS}.

The evaluation was conducted on open-source datasets commonly used for scene reconstruction and localization with NeRF and 3DGS: Synthetic NeRF~\cite{mildenhall2021nerf}, Mip-NeRF360~\cite{barron2022mipnerf360}, and Tanks and Temples~\cite{knapitsch2017tanks}. These datasets contain hundreds of images captured in diverse indoor and outdoor scenes from varying viewpoints, either synthetic or realistic, with each image labeled with ground-truth 3D camera poses.

We chose commonly used evaluation metrics: mean rotation error (RE), mean translation error (TE), the percentage of results with $\rm{RE} <5^{\circ}$ and $\rm{TE} <0.05$, and the mean inference time per image. As scene scale varies across datasets, we normalized TE in each scene by setting the scale to the mean Euclidean distance of all camera poses from their centroid. Doing so is reasonable because all images in each scene were captured around a central object and is also reproducible since ground-truth camera poses are available for all datasets. This setup also enables fair comparisons of TE and of the condition $\rm{TE} < 0.05$ across scenes.

\subsection{Implementation details}

\subsubsection{Scene reconstruction}

Visual localization requires a scene representation (e.g., 3DGS) as well as a query image.
We used gsplat~\cite{ye2025gsplat}, an implementation of Gaussian splatting~\cite{kerbl2023gs} that offers improved reconstruction efficiency and comparable rendering quality, while also supporting depth rendering---a feature not available in the original implementation.
We used NerfBaselines~\cite{kulhanek2024nerfbaselines}, a framework for reconstructing and interfacing with gsplat~\cite{ye2025gsplat}.
% On which subsets we reconstruct and query for visual localization
We used the existing train and test splits from the Synthetic NeRF dataset for scene reconstruction and query images in visual localization.
For the Mip-NeRF360 and Tanks and Temples datasets, we selected every eighth image as a query, using the remaining images for scene reconstruction.

% How scenes are reconstructed.
All baseline methods also require a scene reconstruction, similar to ours.
PiNeRF requires Instant-NGP~\cite{mueller2022instantngp}, a NeRF variant with significantly improved reconstruction speed.
iComMa and 6DGS require the original Gaussian splatting~\cite{kerbl2023gs}.
The reconstructions for all scenes, in each respective representation, were conducted in advance of the visual localization experiments.

\subsubsection{Visual localization}

We used SuperPoint~\cite{detone2018superpoint} and SuperGlue~\cite{sarlin2020superglue}, a learning-based feature point extractor and matcher, which shows robust performance compared to traditional methods (e.g., SIFT~\cite{lowe2004sift}), with comparable runtime when run on GPU.
We used the pre-trained model trained on its indoor datasets and the default values for keypoint and match thresholds (0.005 and 0.2, respectively), without any additional parameter tuning.
For pose estimation, we applied PnP with 50 iterations, followed by RANSAC with a a reprojection error threshold corresponding to approximately 1\% of the image width.
If the number of 2D–3D correspondences was insufficient for PnP to proceed, or if iterative PnP failed to converge, our method returned the initial pose estimate as the final result.

We used the default setups and hyperparameters for the baseline methods---piNeRF, iComMa, and 6DGS---without any additional tuning.

% Computing setup
All experiments---including scene reconstruction using either NeRF or 3DGS and the subsequent visual localization---were conducted on a system equipped with a 32-core 13th Gen Intel(R) Core(TM) i9-13900K CPU (up to 5.8 GHz) and a single NVIDIA GeForce RTX 4090 GPU.

\subsection{Comparison under randomized initial poses}
\label{subsection:Comparison under randomized initial poses}

\begin{table*}[!htbp]
    \vspace{1.5mm}
    \centering\caption{\textsc{Comparison of visual localization methods under randomized initial poses}}
    \resizebox{0.9\textwidth}{!}{%    
    \begin{tabular}{l|ccc|ccc|ccc}
        \toprule
         & \multicolumn{3}{c|}{\textbf{Synthetic NeRF}} & \multicolumn{3}{c|}{\textbf{Mip-NeRF360}} & \multicolumn{3}{c}{\textbf{Tanks and Temples}} \\
         & piNeRF & iComMa & Ours & piNeRF & iComMa & Ours & piNeRF & iComMa & Ours \\ \midrule
        RE ($^{\circ}$) & 3.08 & 13.29 & \textbf{1.61} & 14.36 & 2.67 & \textbf{1.63} & 21.27 & 31.71 & \textbf{11.10} \\ 
        TE (unitless) & 0.05 & 0.20 & \textbf{0.03} & 0.25 & \textbf{0.04} & \textbf{0.04} & 0.27 & 0.38 & \textbf{0.20} \\ 
        $\textrm{RE}<5^{\circ},~\textrm{TE}<0.05$ (\%) & 71.88 & 79.00 & \textbf{90.94} & 1.22 & 86.99 & \textbf{90.65} & 6.89 & 32.45 & \textbf{42.42} \\ 
        Time / Image (s) & 13.56 & 16.74 & \textbf{0.09} & 13.46 & 27.78 & \textbf{0.15} & 14.59 & 41.79 & \textbf{0.10} \\ \bottomrule
    \end{tabular}
    \label{table:results-random-aggregated}
    }
    \captionsetup{justification=raggedright}
    \caption*{\footnotesize \textit{* Synthetic NeRF (8 scenes, 1600 test images), Mip-NeRF360 (9 scenes, 221 test images), and Tanks and Temples (21 scenes, 943 test images); TE is normalized by the trajectory radius for each scene.
    % * RE: rotation error ($^{\circ}$); TE: translation error (unitless); $\textrm{RE}<5^{\circ}~\textrm{TE}<0.05$: percentage of results with RE $<5^{\circ}$ and TE $<0.05$; Time: time per image (s). \\
    }}
\end{table*}

\begin{table}[!htbp]
    \centering\caption{\textsc{Comparison of our method with Chen \textit{et al.}~\cite{chen2024nerffeaturmatching} on Synthetic NeRF dataset}}
    \resizebox{0.75\linewidth}{!}{%    
    \begin{tabular}{l|cccc}
        \toprule
         & \makecell{Chen\\(Full)} & \makecell{Chen\\(Lite)} & \makecell{Ours\\(Avg.)} & \makecell{Ours\\(Best)} \\ \midrule
        RE ($^{\circ}$) & 1.25 & 1.57 & 4.77 & 0.19 \\ 
        TE (unitless) & 0.02 & 0.02 & 0.02 & 0.00 \\ 
        $\textrm{TE}<0.05$ (\%) & 95.00 & 94.50 & 87.78 & 99.69 \\ 
        $\textrm{RE}<5^{\circ}$ (\%) & 88.00 & 75.00 & 96.71 & 99.81 \\ 
        \bottomrule
    \end{tabular}
    \label{table:results-chen}
    }
    \captionsetup{justification=raggedright}
    \caption*{\footnotesize \textit{* Chen (Full): Chen \textit{et al.}’s full pipeline; Chen (Lite): Without post-pose refinement; Ours (Avg.): Our method, averaging the results of 5 random initial poses per test image; Ours (Best): Our method, reporting the best result among 5 random initial poses per test image.}
    % * RE: rotation error ($^{\circ}$); TE: translation error (unitless); $\textrm{RE}<5^{\circ}~\textrm{TE}<0.05$: percentage of results with RE $<5^{\circ}$ and TE $<0.05$; Time: time per image (s). \\
    }
\end{table}

First, we compare the results when the initial pose of the query image is generated by applying rotations and translations to the ground-truth pose, with rotation axes and translation directions sampled uniformly at random and magnitudes sampled from zero-mean normal distributions such that 95\% of the samples fall within $\Delta \theta$ and $\Delta p$, respectively.
% This setup is reasonable, as all the methods we compare---iNeRF~\cite{lin2023pnerf}, iComMa~\cite{sun2023icomma}, and our method---require initial pose estimates for the query images.
Here, $\Delta \theta$ and $\Delta p$ were determined based on the camera’s field of view, the radius of the camera trajectory in each scene, and the assumption that the centered object has a radius equal to half of the trajectory radius, such that the deviation results in the object leaving the image frame.
This level of deviation reflects the typical roughness of initial pose estimates (e.g., from image retrieval, auxiliary sensors, or other methods), and aligns with the operating conditions of the methods being compared.

Table~\ref{table:results-random-aggregated} shows the comparison between our method and the baseline methods.
The aggregated results on the Synthetic NeRF, Mip-NeRF360, and Tanks and Temples datasets, are reported.
% We report the mean rotation error (RE), mean translation error (TE), the percentage of results with $\rm{RE} < 5^{\circ}$ and $\rm{TE} < 0.05$, and the mean inference time per image.
% Translation errors were normalized per scene by setting the scene scale to the camera trajectory's radius.
% The method achieving the best performance in each metric for the aggregated results on each dataset is highlighted in bold.
Our method consistently achieves the best accuracy and efficiency across all datasets.
Particularly, compared to iComMa---a 3DGS-based visual localization method by minimizing photometric loss between the query image and a rendered image at the estimated pose---our method shows up to about 12\% improvement in $\rm{RE} < 5^{\circ}$ and $\rm{TE} < 0.05$ on the Synthetic NeRF dataset, while also reducing inference time per image by more than two orders of magnitude---from more than 10 seconds to as fast as 0.1 seconds across all datasets.

\subsection{Comparison with a method based on feature correspondence using NeRF}
\label{subsection:Comparison with a method based on feature correspondence using NeRF}

We provide a comparison with Chen \textit{et al.}~\cite{chen2024nerffeaturmatching}, a method based on feature correspondence using NeRF as a scene representation---part of which our method uses---in this separate subsection, as their open-source implementation is not available and their experimental setup does not align with that in Section~\ref{subsection:Comparison under randomized initial poses}.
We use the exact numbers reported in their paper, evaluated on the Synthetic NeRF dataset, with only the translation error normalized by scene scale to enable a fair comparison with ours (Table~\ref{table:results-chen}).

Chen (Full) refers to the results from their full pipeline, while Chen (Lite) uses the same setup but without post-pose refinement.
Chen reports results using only five randomly selected test images per scene (out of 200), each evaluated under five random initial poses, generated by applying a translation uniformly sampled between 0 and 0.2 along a random direction, followed by a rotation with a magnitude uniformly sampled between 10$^{\circ}$ and 40$^{\circ}$ around a random axis.
However, it is not clearly stated whether their reported results show an average over the five initial poses or the best among them.

To ensure a fair comparison, we used the same protocol for generating initial poses as described by Chen.
Due to the ambiguity regarding whether their results show an average or the best case, we report both: the results averaged over five random initial poses per test image (Ours (Avg.)) and the best results among the five in rotation error (Ours (Best)).
Additionally, while Chen evaluates only five test images per scene, our method is evaluated on all 200 test images per scene, totaling 1,600 evaluations across all scenes.

% Table~\ref{table:results-chen} shows that our method (Ours (Best)) outperforms both Chen (Full) and Chen (Lite) across all reported metrics: rotation error (RE), translation error (TE), percentage of results with $\textrm{RE} < 5^{\circ}$, and percentage with $\textrm{TE} < 5$.

\subsection{Comparison under initial poses provided by 6DGS}
\label{subsection:Comparison under initial poses provided by 6DGS}

\begin{table*}[!htbp]
    \vspace{1.5mm}
    \centering\caption{\textsc{Comparison of visual localization methods under initial poses provided by 6DGS}}
    \resizebox{1.0\textwidth}{!}{%    
    \begin{tabular}{l|c|ccc|c|ccc|c|ccc}
        \toprule
         & \multicolumn{4}{c|}{\textbf{Synthetic NeRF}} & \multicolumn{4}{c|}{\textbf{Mip-NeRF360}} & \multicolumn{4}{c}{\textbf{Tanks and Temples}} \\
         & 6DGS & piNeRF & iComMa & Ours & 6DGS & piNeRF & iComMa & Ours & 6DGS & piNeRF & iComMa & Ours \\ \midrule
        RE ($^{\circ}$) & 17.93 & 23.69 & 35.61 & \textbf{14.09} & 20.84 & 17.04 & 9.60 & \textbf{6.47} & 17.93 & 27.96 & 35.69 & \textbf{11.21} \\ 
        TE (unitless) & 0.59 & 0.52 & 0.51 & \textbf{0.27} & 0.22 & 0.28 & 0.11 & \textbf{0.05} & 0.59 & 0.23 & 0.39 & \textbf{0.08} \\ 
        $\textrm{RE}<5^{\circ},~\textrm{TE}<0.05$ (\%) & 0.06 & 10.44 & 44.25 & \textbf{61.25} & 0.41 & 2.44 & 78.46 & \textbf{90.24} & 0.06 & 8.91 & 31.60 & \textbf{72.53} \\ 
        Time / Image (s) & 0.04 & 14.62 & 29.92 & \textbf{0.09} & 0.02 & 14.18 & 30.65 & \textbf{0.15} & 0.04 & 14.88 & 45.68 & \textbf{0.11} \\ 
        \bottomrule
    \end{tabular}
    \label{table:results-6dgs-aggregated}
    }
    \captionsetup{justification=raggedright}
    \caption*{\footnotesize \textit{* Synthetic NeRF (8 scenes, 1600 test images), Mip-NeRF360 (9 scenes, 221 test images), and Tanks and Temples (21 scenes, 943 test images); TE is normalized by the trajectory radius for each scene; 6DGS results (which do not require an initial pose) are included for reference.
    % * RE: rotation error ($^{\circ}$); TE: translation error (unitless); $\textrm{RE}<5^{\circ}~\textrm{TE}<0.05$: percentage of results with RE $<5^{\circ}$ and TE $<0.05$; Time: time per image (s). \\
    }}
\end{table*}

\begin{figure*}[!htbp]
    \centering
    \captionsetup[subfloat]{labelformat=parens}

    % \includegraphics[width=0.275\textwidth]{imgs/mipnerf360_garden_DSC08060_query.png}%
    % \quad
    % \includegraphics[width=0.275\textwidth]{imgs/mipnerf360_garden_DSC08060_6dgs.png}%
    % \quad
    % \includegraphics[width=0.275\textwidth]{imgs/mipnerf360_garden_DSC08060_ours.png}%

    % \vspace{0.75em}

    % % Bottom row: with subfloat labels
    % \subfloat[Query]{\includegraphics[width=0.275\textwidth]{imgs/tanksandtemples_train_00081_query.png}}%
    % \quad
    % \subfloat[Initial Pose]{\includegraphics[width=0.275\textwidth]{imgs/tanksandtemples_train_00081_6dgs.png}}%
    % \quad
    % \subfloat[Estimate]{\includegraphics[width=0.275\textwidth]{imgs/tanksandtemples_train_00081_ours.png}}%

    \includegraphics[width=0.48\textwidth]{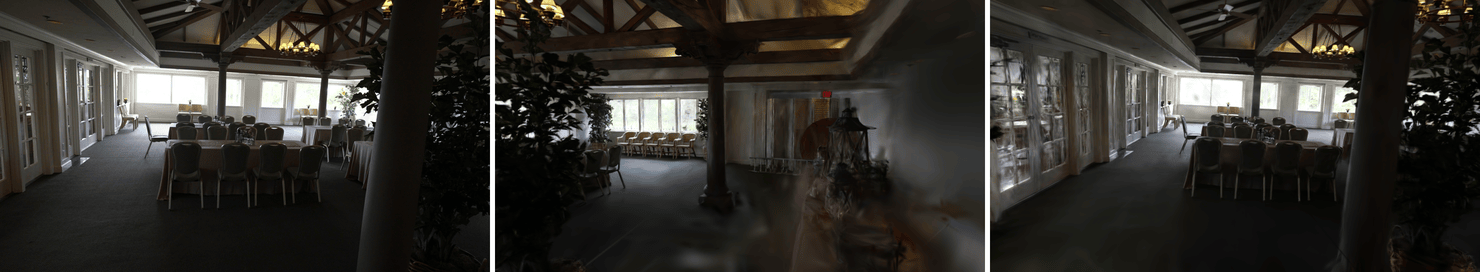}%
    \quad
    \includegraphics[width=0.48\textwidth]{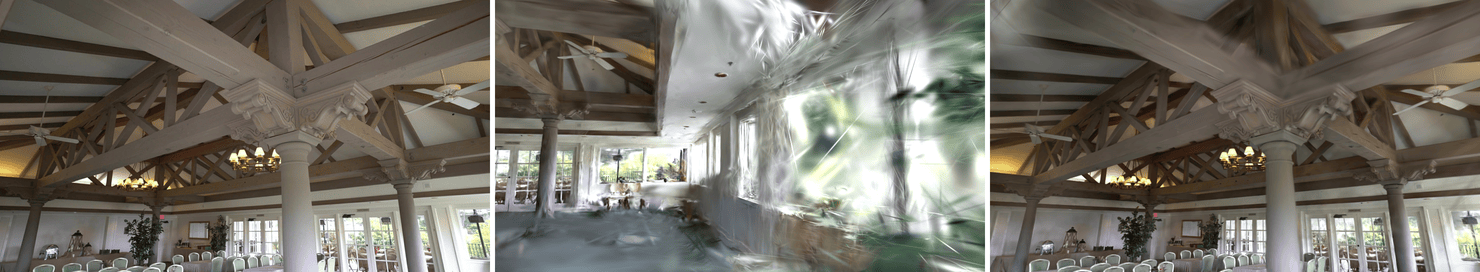}%
    \vspace{0.75em}
    
    \includegraphics[width=0.48\textwidth]{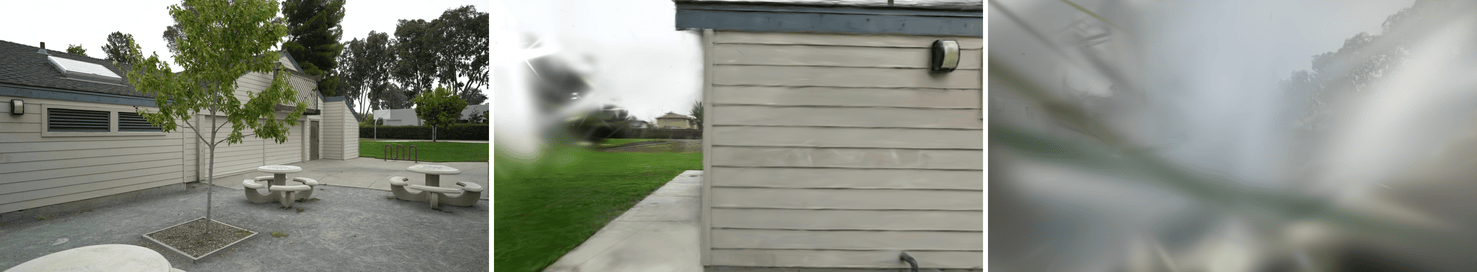}%
    \quad
    \includegraphics[width=0.48\textwidth]{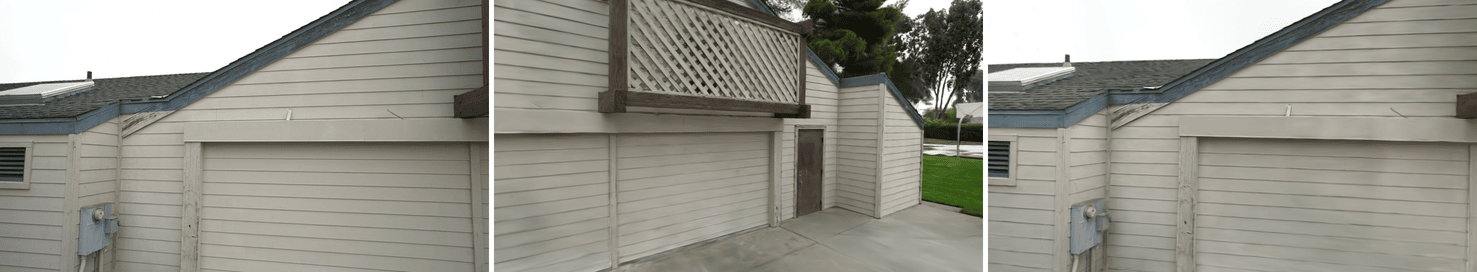}%
    \vspace{0.75em}
    
    \includegraphics[width=0.48\textwidth]{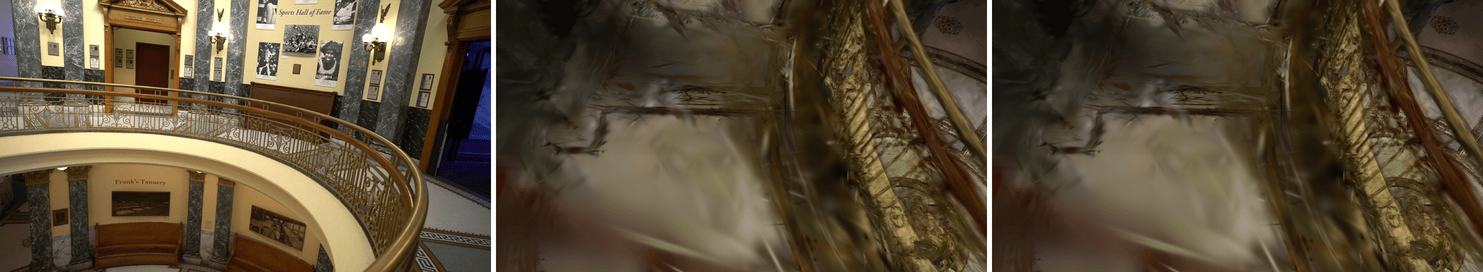}%
    \quad
    \includegraphics[width=0.48\textwidth]{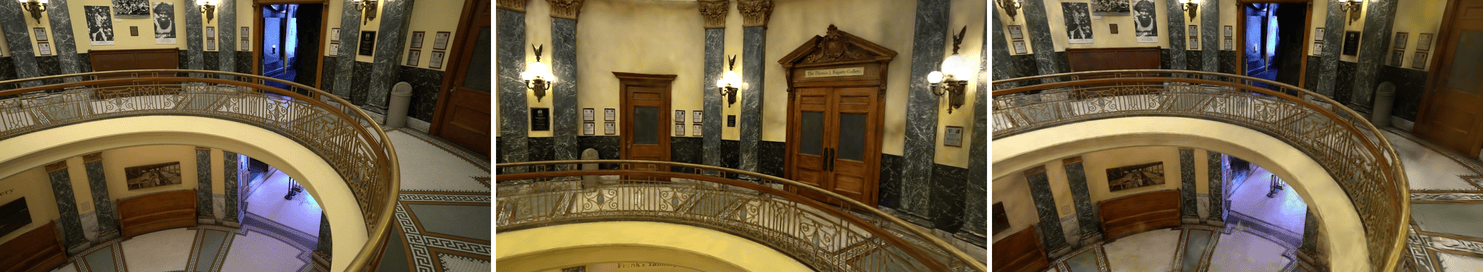}%
    \vspace{0.75em}

    % Bottom row: with subfloat labels
    \subfloat[Failure ($\rm{RE} \geq 5^{\circ}$ or $\rm{TE} \geq 0.05$)]{\includegraphics[width=0.48\textwidth]{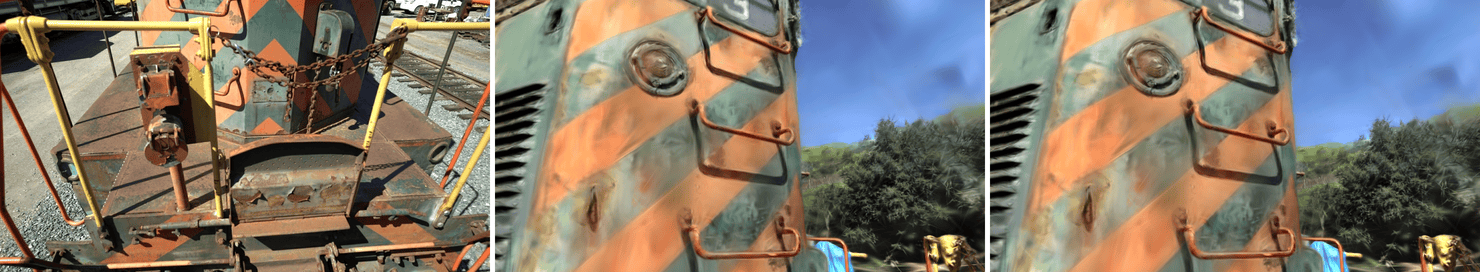}}%
    \quad
    \subfloat[Success ($\rm{RE} < 5^{\circ}$ and $\rm{TE} < 0.05$)]{\includegraphics[width=0.48\textwidth]{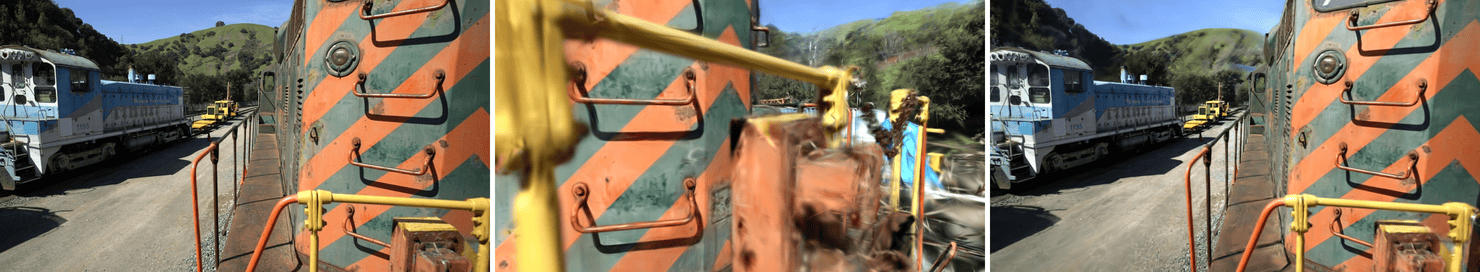}}%

    \caption{
    Example pairs of failure and success cases from several scenes in the Tanks and Temples dataset, where success is defined as $\mathrm{RE} < 5^{\circ}$ and $\mathrm{TE} < 0.05$.
    Each triplet shows, from left to right, the query image, the image rendered at the initial pose provided by 6DGS, and the image rendered at the pose estimated by our method.
    }
    \label{fig:results-tanksandtemples}
\end{figure*}

Next, we compare the results when the initial pose of the query image is provided by 6DGS~\cite{matteo2024-6dgs}. This setup is reasonable, as 6DGS can estimate rough poses (e.g., approximately $20^{\circ}$ in rotation error and 0.2-0.5 in normalized translation error), and can therefore serve as a source of initial poses for the methods we compare.

Table~\ref{table:results-6dgs-aggregated} shows comparisons between our method and baseline methods on the three datasets. 
% In particular, we compare results when the pose estimates from 6DGS are used as initial guesses for piNeRF, iComMa, and our method.
The original 6DGS results---which do not require any initial pose but do require a model pretrained on each scene as a prior step---are included as a reference in the leftmost subcolumn of each dataset column.
% (similar to how other methods require reconstructed scenes as a prerequisite, with 6DGS itself relying on Gaussian splatting as well)
% We report the mean rotation error (RE), mean translation error (TE), the percentage of results with $\rm{RE} < 5^{\circ}$ and $\rm{TE} < 0.05$, and the mean inference time per image.
% Translation errors were normalized per scene by setting the scene scale to the camera trajectory's radius.
% The method achieving the best performance in each metric for the aggregated results on each dataset is highlighted in bold.
Our method consistently achieves the best accuracy and efficiency across all datasets.
Particularly, compared to iComMa, our method shows over a 40\% improvement in $\rm{RE} < 5^{\circ}$ and $\rm{TE} < 0.05$ on the Tanks and Temples dataset, while also reducing inference time per image by more than two orders of magnitude---from more than 10 seconds to as fast as 0.1 seconds across all datasets.

Figure~\ref{fig:results-tanksandtemples} shows examples of both failure and success cases from our method on several scenes from the Tanks and Temples dataset, where success is defined as $\mathrm{RE} < 5^{\circ}$ and $\mathrm{TE} < 0.05$.
Each triplet shows, from left to right, the query image, the image rendered at the initial pose provided by 6DGS, and the image rendered at the pose estimated by our method.
Failures (Fig.~\ref{fig:results-tanksandtemples}(a)) occur either due to inaccurate pose estimation despite feature matching (top two rows), or due to PnP failure caused by insufficient feature matches between the query and the rendered image---resulting in the initial pose being returned (bottom two rows).
Nonetheless, even within the same scene, our method shows success cases as well (Fig.~\ref{fig:results-tanksandtemples}(b)) despite drastic appearance differences between the query and the rendered image.
% Again, our method performs only a single rendering per query image---at the initial pose, as shown in column (b) of Fig.~\ref{fig:results-tanksandtemples}---which significantly contributes to its inference time improvement over the baselines.

\subsection{Sensitivity to error in the initial pose estimate}
\label{subsection:Sensitivity to initial pose from ground-truth}

\begin{figure}[!htbp]
    \centering
    \includegraphics[width=0.9\linewidth, keepaspectratio]{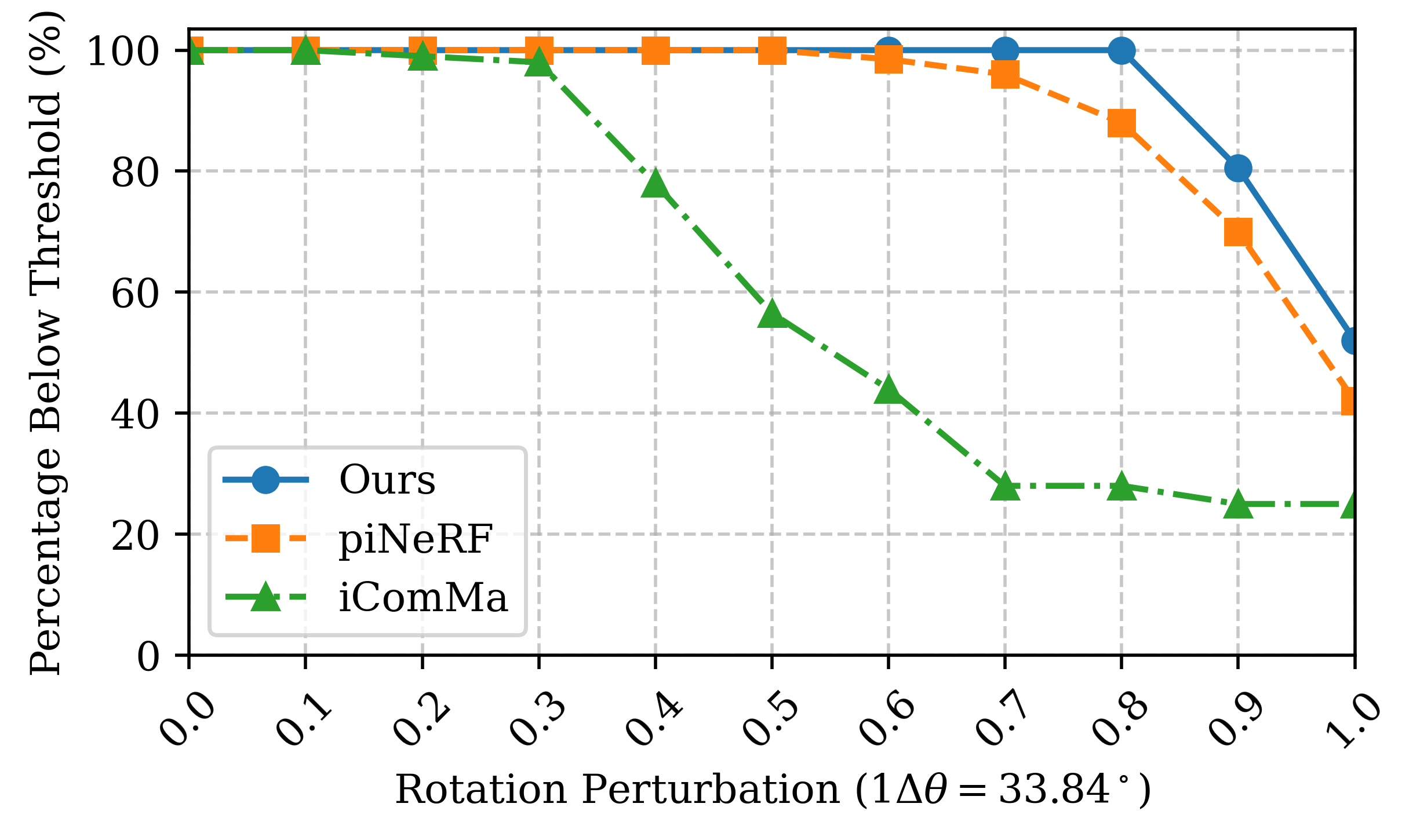}
    \caption{Percentage of pose estimates (out of 200 test images) with rotation error $<5^{\circ}$ and normalized translation error $<0.05$ as a function of difference between the initial pose estimate and ground-truth in yaw on the Lego scene (unit: \%). $\Delta \theta = \rm{33.84}^{\circ}$ is determined as the minimum yaw error required for the object to move out of view.}
    \label{fig:sensitivity-rotation}
\end{figure}

\begin{figure}[!htbp]
    \centering
    \includegraphics[width=0.9\linewidth, keepaspectratio]{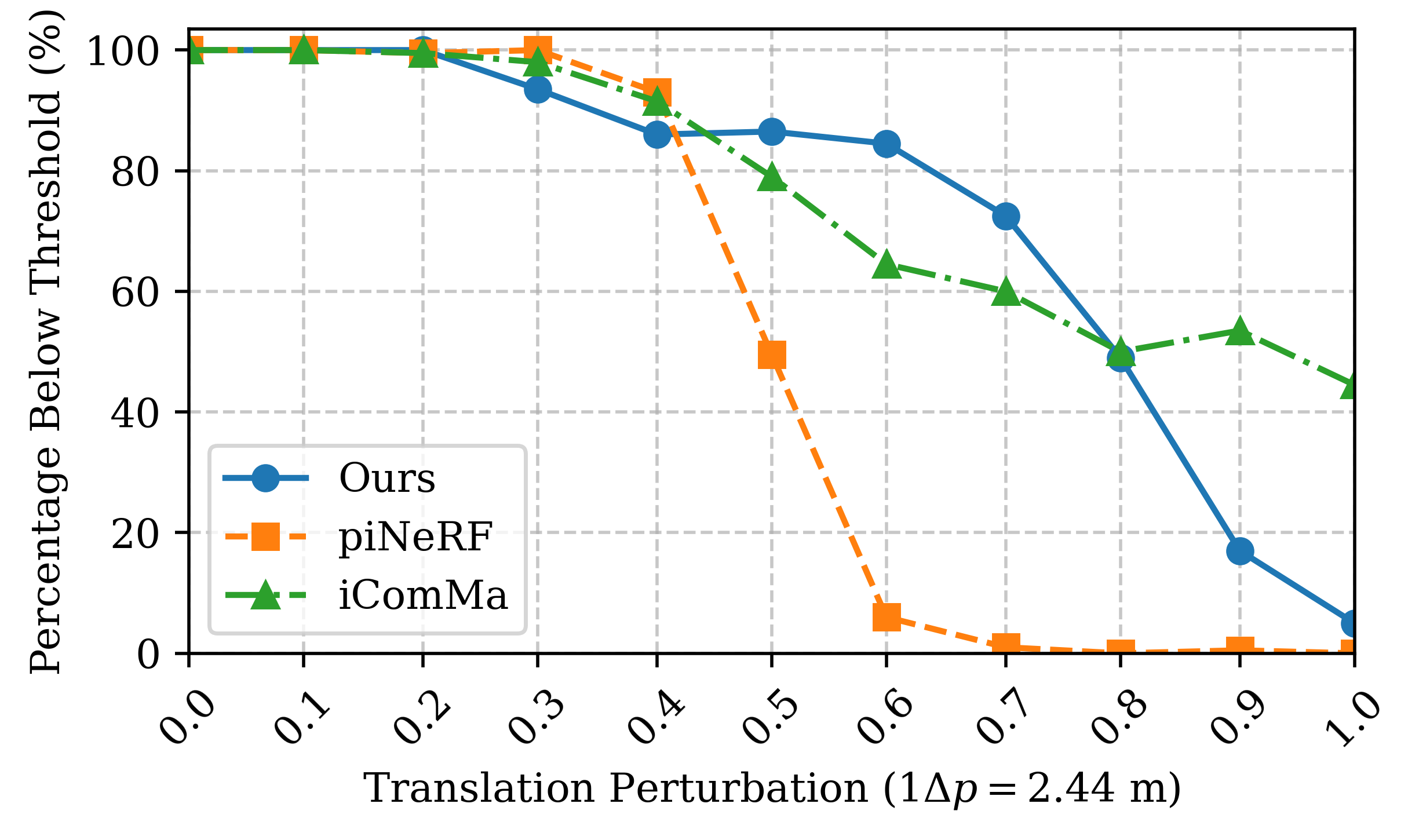}
    \caption{Percentage of pose estimates (out of 200 test images) with rotation error $<5^{\circ}$ and normalized translation error $<0.05$ as a function of difference between the initial pose and ground-truth in x-position on the Lego scene (unit: \%). $\Delta p = \rm{2.44}~\rm{m}$ is determined as the minimum translation error required for the object to move out of view.}
    \label{fig:sensitivity-translation}
\end{figure}

We also study the extent to which our method can reliably estimate the pose of a query image
despite error in the initial pose estimate.
% when its initial pose deviates from the ground truth.
% We compare piNeRF~\cite{lin2023pnerf}, iComMa~\cite{sun2023icomma}, and our approach, all of which require an initial pose estimate for each query image.
We report the percentage of results with rotation and translation (normalized by scene scale) errors below threshold (i.e., $\textrm{RE}<5^{\circ},~\textrm{TE}< 0.05$)
as a function of error in the initial pose estimate.
%, under varying degrees of initial pose deviated from the ground-truth.
Specifically, we examine cases where the initial rotation error is between 0 and $1\Delta \theta$ in yaw, and the initial translation error is between 0 and $\Delta p$ along x-axis.
Here, $\Delta \theta$ and $\Delta p$ were determined based on the camera’s field of view, the radius of the camera trajectory in each scene, and the assumption that the centered object has a radius equal to half of the trajectory radius, such that the deviation would result in the object leaving the image frame.

Figures~\ref{fig:sensitivity-rotation} and~\ref{fig:sensitivity-translation} show results for the Lego scene in the Synthetic NeRF dataset, where $\Delta \theta$ and $\Delta p$ are $\rm{33.84}^{\circ}$ and $\rm{2.44}~\mathrm{m}$, respectively.
Our method achieves the best performance up to $1.0 \Delta \theta$ in rotation and $0.8 \Delta p$ in translation among the three methods.
% Notably, it shows significantly higher success rates in rotation --- 100\% even at $0.8 \Delta \theta$, where iComMa and piNeRF show only 28\% and 0\%, respectively.

\subsection{Choice of feature point extractor and matcher}
\label{subsection:Choice of feature point extractor and matcher}

\begin{table}[!tbp]
    \centering
    \vspace{1.5mm}
    \caption{\textsc{Comparison of feature extractor and matcher choices on three datasets}}
    \resizebox{0.9\linewidth}{!}{%
    \begin{tabular}{ll|cccc}
    \toprule
        ~ & ~ & RE & TE & Acc & Time \\ \midrule
        \multirow{3}{*}{Synthetic NeRF} & SIFT & 7.04 & 0.37 & 66.94 & 0.14 \\ 
        ~ & SP+SG & \textbf{1.61} & \textbf{0.03} & 90.94 & \textbf{0.09} \\ 
        ~ & LoFTR & 1.84 & \textbf{0.03} & \textbf{91.94} & 0.13 \\  \midrule
        \multirow{3}{*}{Mip-NeRF360} & SIFT & \textbf{1.44} & 0.05 & 87.80 & 0.21 \\ 
        ~ & SP+SG & 1.63 & \textbf{0.04} & \textbf{90.65} & \textbf{0.15} \\ 
        ~ & LoFTR & 1.47 & \textbf{0.04} & 90.24 & 0.18 \\ \midrule
        \multirow{3}{*}{Tanks and Temples} & SIFT & 11.85 & 0.59 & 34.47 & 0.14 \\ 
        ~ & SP+SG & \textbf{11.10} & \textbf{0.20} & \textbf{42.42} & \textbf{0.10} \\ 
        ~ & LoFTR & 12.27 & 0.21 & 41.68 & 0.13 \\  \bottomrule
    \end{tabular}
    \label{table:results-feature}
    }
    \captionsetup{justification=raggedright}
    \caption*{\footnotesize \textit{* RE: rotation error ($^{\circ}$); TE: translation error (unitless); Acc: percentage of results with RE $<5^{\circ}$ and TE $<0.05$; Time: time per image (s). \\*** TE is normalized by the trajectory radius for each scene.
    }}
\end{table}

Lastly, we study how the choice of feature extractor and matcher influences the performance of the proposed approach. We compare SIFT~\cite{lowe2004sift}, a commonly used traditional method; SuperPoint~\cite{detone2018superpoint} and SuperGlue~\cite{sarlin2020superglue} (SP+SG), a learning-based method used as our default; and LoFTR~\cite{sun2021loftr}, another newer learning-based method that has been reported to outperform SP+SG.
For SIFT, we applied Lowe’s ratio test~\cite{lowe2004sift} with a threshold of 0.7 to filter out outlier matches. For LoFTR, we used its official pre-trained model trained on the outdoor dataset, without any additional parameter tuning.

Table~\ref{table:results-feature} shows the results of how the choice of feature extractor and matcher affects performance, aggregated over each of the three datasets. 
% We report the mean rotation error (RE), mean translation error (TE), the percentage of results with $\mathrm{RE} < 5^{\circ}$ and $\mathrm{TE} < 0.05$, and the mean inference time per image.
The initial pose estimate for each query image was generated by applying random rotations and translations to the ground-truth pose, with magnitudes sampled from normal distributions such that 95\% of the samples fall within $\Delta \theta$ and $\Delta p$, where $\Delta \theta$ and $\Delta p$ are scene-specific constants. Rotation axes and translation directions were sampled uniformly at random. These initial conditions are the same as those used in Section~\ref{subsection:Comparison under randomized initial poses}.
The results show that learning-based methods---SP+SG and LoFTR---consistently outperform SIFT in terms of TE, RE, and the percentage of results with $\mathrm{RE} < 5^{\circ}$ and $\mathrm{TE} < 0.05$, achieving improvements of over 20\% on the Synthetic NeRF dataset. This is obtained with on par or even better inference time per image.
Between SP+SG and LoFTR, the error metrics are comparable, but SP+SG consistently shows lower inference time. Based on this, we choose SP+SG as the default feature extractor and matcher in our pipeline.

%%%%%%%%%%%%%%%%%%%%%%%%%%%%%%%%%%%%%%%%%%%%%%%%%%%%%%%%%%%%%%%%%%%%%%%%%%%%%%

\section{Conclusion}
\label{section:Conclusion}

In this paper, we proposed a visual localization approach using 3DGS as a scene representation.
% Our method follows a conventional pipeline. It establishes feature correspondences between the query image and the image rendered at the initial pose estimate, followed by applying perspective-n-point (PnP) using a depth map rendered from the 3DGS representation.
We showed that our method outperforms existing approaches that estimate camera poses by minimizing the photometric difference between the query image and the rendered image at the estimated pose, either using NeRF and 3DGS as a scene representation. Our method yielded an improvement of more than two orders of magnitude in inference time along with significant gains in accuracy and better tolerance of errors in the initial pose estimate.
%a wider range of initial poses deviating from the ground-truth.

\subsection{Limitations of our method and of baseline methods}

% Once a reasonable range of initial pose estimates is given---specific to each method (ours and the baselines)---
The performance of our method and of the baseline methods depends on the quality of the reconstructed scene, whether represented by 3DGS or NeRF. High-quality reconstruction yields better rendered images, which are essential for methods using feature correspondence or photometric loss minimization.
Additionally, our method depends on the accuracy of rendered depth maps for correct PnP results, which, again, relies on scene quality.

\subsection{Future work}

One direction for future work is to improve the robustness of our method. For instance, a failure mode may occur when the number of feature correspondences is too low (e.g., fewer than 10) to yield a reliable pose estimate. To address this, one could render an intermediate image from the pose estimate and perform feature matching between this image and the query image to obtain the final pose. This additional step may help eliminate such failure cases by increasing the number of valid correspondences between the rendered image at the intermediate pose and the query image, at the cost of a few extra iterations and slightly increased inference time.

Another direction is to integrate our method into a broader robotics pipeline.
For example, most visual SLAM systems using 3DGS representations (e.g., Gaussian Splatting SLAM~\cite{matsuki2024gsslam}) rely on photometric loss minimization to track incoming frames. While this approach works well when frame-to-frame motion is small, it may fail when the motion becomes large.
Our method could serve as an alternative module in such scenarios, helping to accurately localize incoming frames that undergo abrupt motion.
This role is analogous to relocalization in conventional visual SLAM systems (e.g., ORB-SLAM3~\cite{campos2021orbslam3}) when tracking is lost.

A third possible direction is to apply our method to real-world robotics tasks. Since our method (as well as the baselines) is evaluated on scenes captured around a central object, one could envision applications---presumably using the same pipeline with little or no modification---where a robot localizes itself while observing the object, assuming a pre-constructed 3DGS scene of that object is available.

%%%%%%%%%%%%%%%%%%%%%%%%%%%%%%%%%%%%%%%%%%%%%%%%%%%%%%%%%%%%%%%%%%%%%%%%%%%%%%
% \appendices

%%%%%%%%%%%%%%%%%%%%%%%%%%%%%%%%%%%%%%%%%%%%%%%%%%%%%%%%%%%%%%%%%%%%%%%%%%%%%%
% \section*{Acknowledgement}
% This work was supported by the NASA Grant No. STTR-80NSSC20C0020.

\balance
\bibliographystyle{IEEEtran}
\bibliography{IEEEabrv,References}

\begin{thebibliography}{10}
\providecommand{\url}[1]{#1}
\csname url@samestyle\endcsname
\providecommand{\newblock}{\relax}
\providecommand{\bibinfo}[2]{#2}
\providecommand{\BIBentrySTDinterwordspacing}{\spaceskip=0pt\relax}
\providecommand{\BIBentryALTinterwordstretchfactor}{4}
\providecommand{\BIBentryALTinterwordspacing}{\spaceskip=\fontdimen2\font plus
\BIBentryALTinterwordstretchfactor\fontdimen3\font minus
  \fontdimen4\font\relax}
\providecommand{\BIBforeignlanguage}[2]{{%
\expandafter\ifx\csname l@#1\endcsname\relax
\typeout{** WARNING: IEEEtran.bst: No hyphenation pattern has been}%
\typeout{** loaded for the language `#1'. Using the pattern for}%
\typeout{** the default language instead.}%
\else
\language=\csname l@#1\endcsname
\fi
#2}}
\providecommand{\BIBdecl}{\relax}
\BIBdecl

\bibitem{miao2024survey-vloc}
J.~Miao, K.~Jiang, T.~Wen, Y.~Wang, P.~Jia, B.~Wijaya, X.~Zhao, Q.~Cheng,
  Z.~Xiao, J.~Huang \emph{et~al.}, ``A survey on monocular re-localization:
  From the perspective of scene map representation,'' \emph{IEEE Transactions
  on Intelligent Vehicles}, 2024.

\bibitem{kerbl2023gs}
B.~Kerbl, G.~Kopanas, T.~Leimk{\"u}hler, and G.~Drettakis, ``3d gaussian
  splatting for real-time radiance field rendering.'' \emph{ACM Trans. Graph.},
  vol.~42, no.~4, pp. 139--1, 2023.

\bibitem{mildenhall2021nerf}
B.~Mildenhall, P.~P. Srinivasan, M.~Tancik, J.~T. Barron, R.~Ramamoorthi, and
  R.~Ng, ``Nerf: Representing scenes as neural radiance fields for view
  synthesis,'' \emph{Communications of the ACM}, vol.~65, no.~1, pp. 99--106,
  2021.

\bibitem{zhu2024survey-3dgs-robotics}
S.~Zhu, G.~Wang, D.~Kong, and H.~Wang, ``3d gaussian splatting in robotics: A
  survey,'' \emph{arXiv preprint arXiv:2410.12262}, 2024.

\bibitem{sun2023icomma}
Y.~Sun, X.~Wang, Y.~Zhang, J.~Zhang, C.~Jiang, Y.~Guo, and F.~Wang, ``icomma:
  Inverting 3d gaussians splatting for camera pose estimation via comparing and
  matching,'' \emph{arXiv preprint arXiv:2312.09031}, 2023.

\bibitem{jiang2024gsreloc}
P.~Jiang, G.~Pandey, and S.~Saripalli, ``3dgs-reloc: 3d gaussian splatting for
  map representation and visual relocalization,'' \emph{arXiv preprint
  arXiv:2403.11367}, 2024.

\bibitem{botashev2024gsloc}
K.~Botashev, V.~Pyatov, G.~Ferrer, and S.~Lefkimmiatis, ``Gsloc: Visual
  localization with 3d gaussian splatting,'' in \emph{2024 IEEE/RSJ
  International Conference on Intelligent Robots and Systems (IROS)}.\hskip 1em
  plus 0.5em minus 0.4em\relax IEEE, 2024, pp. 5664--5671.

\bibitem{jun2024gsloc}
H.~Jun, H.~Yu, and S.~Oh, ``Renderable street view map-based localization:
  Leveraging 3d gaussian splatting for street-level positioning,'' in
  \emph{2024 IEEE/RSJ International Conference on Intelligent Robots and
  Systems (IROS)}.\hskip 1em plus 0.5em minus 0.4em\relax IEEE, 2024, pp.
  5635--5640.

\bibitem{yen2021inerf}
L.~Yen-Chen, P.~Florence, J.~T. Barron, A.~Rodriguez, P.~Isola, and T.-Y. Lin,
  ``inerf: Inverting neural radiance fields for pose estimation,'' in
  \emph{2021 IEEE/RSJ International Conference on Intelligent Robots and
  Systems (IROS)}.\hskip 1em plus 0.5em minus 0.4em\relax IEEE, 2021, pp.
  1323--1330.

\bibitem{chen2024nerffeaturmatching}
R.~Chen, Y.~Cong, and Y.~Ren, ``Marrying nerf with feature matching for
  one-step pose estimation,'' in \emph{2024 IEEE International Conference on
  Robotics and Automation (ICRA)}, 2024, pp. 7302--7309.

\bibitem{lin2023pnerf}
Y.~Lin, T.~M{\"u}ller, J.~Tremblay, B.~Wen, S.~Tyree, A.~Evans, P.~A. Vela, and
  S.~Birchfield, ``Parallel inversion of neural radiance fields for robust pose
  estimation,'' in \emph{2023 IEEE International Conference on Robotics and
  Automation (ICRA)}.\hskip 1em plus 0.5em minus 0.4em\relax IEEE, 2023, pp.
  9377--9384.

\bibitem{matteo2024-6dgs}
B.~Matteo, T.~Tsesmelis, S.~James, F.~Poiesi, and A.~Del~Bue, ``6dgs: 6d pose
  estimation from a single image and a 3d gaussian splatting model,'' in
  \emph{European Conference on Computer Vision}.\hskip 1em plus 0.5em minus
  0.4em\relax Springer, 2024, pp. 420--436.

\bibitem{madsen2004lm}
K.~Madsen, H.~B. Nielsen, and O.~Tingleff, ``Methods for non-linear least
  squares problems,'' 2004.

\bibitem{barron2022mipnerf360}
J.~T. Barron, B.~Mildenhall, D.~Verbin, P.~P. Srinivasan, and P.~Hedman,
  ``Mip-nerf 360: Unbounded anti-aliased neural radiance fields,'' in
  \emph{Proceedings of the IEEE/CVF conference on computer vision and pattern
  recognition}, 2022, pp. 5470--5479.

\bibitem{knapitsch2017tanks}
A.~Knapitsch, J.~Park, Q.-Y. Zhou, and V.~Koltun, ``Tanks and temples:
  Benchmarking large-scale scene reconstruction,'' \emph{ACM Transactions on
  Graphics (ToG)}, vol.~36, no.~4, pp. 1--13, 2017.

\bibitem{ye2025gsplat}
V.~Ye, R.~Li, J.~Kerr, M.~Turkulainen, B.~Yi, Z.~Pan, O.~Seiskari, J.~Ye,
  J.~Hu, M.~Tancik \emph{et~al.}, ``gsplat: An open-source library for gaussian
  splatting,'' \emph{Journal of Machine Learning Research}, vol.~26, no.~34,
  pp. 1--17, 2025.

\bibitem{kulhanek2024nerfbaselines}
J.~Kulhanek and T.~Sattler, ``Nerfbaselines: Consistent and reproducible
  evaluation of novel view synthesis methods,'' \emph{arXiv preprint
  arXiv:2406.17345}, 2024.

\bibitem{mueller2022instantngp}
\BIBentryALTinterwordspacing
T.~M\"uller, A.~Evans, C.~Schied, and A.~Keller, ``Instant neural graphics
  primitives with a multiresolution hash encoding,'' \emph{ACM Trans. Graph.},
  vol.~41, no.~4, pp. 102:1--102:15, Jul. 2022. [Online]. Available:
  \url{https://doi.org/10.1145/3528223.3530127}
\BIBentrySTDinterwordspacing

\bibitem{detone2018superpoint}
D.~DeTone, T.~Malisiewicz, and A.~Rabinovich, ``Superpoint: Self-supervised
  interest point detection and description,'' in \emph{Proceedings of the IEEE
  conference on computer vision and pattern recognition workshops}, 2018, pp.
  224--236.

\bibitem{sarlin2020superglue}
P.-E. Sarlin, D.~DeTone, T.~Malisiewicz, and A.~Rabinovich, ``Superglue:
  Learning feature matching with graph neural networks,'' in \emph{Proceedings
  of the IEEE/CVF conference on computer vision and pattern recognition}, 2020,
  pp. 4938--4947.

\bibitem{lowe2004sift}
D.~G. Lowe, ``Distinctive image features from scale-invariant keypoints,''
  \emph{International journal of computer vision}, vol.~60, pp. 91--110, 2004.

\bibitem{sun2021loftr}
J.~Sun, Z.~Shen, Y.~Wang, H.~Bao, and X.~Zhou, ``Loftr: Detector-free local
  feature matching with transformers,'' in \emph{Proceedings of the IEEE/CVF
  conference on computer vision and pattern recognition}, 2021, pp. 8922--8931.

\bibitem{matsuki2024gsslam}
H.~Matsuki, R.~Murai, P.~H. Kelly, and A.~J. Davison, ``Gaussian splatting
  slam,'' in \emph{Proceedings of the IEEE/CVF Conference on Computer Vision
  and Pattern Recognition}, 2024, pp. 18\,039--18\,048.

\bibitem{campos2021orbslam3}
C.~Campos, R.~Elvira, J.~J.~G. Rodr{\'\i}guez, J.~M. Montiel, and J.~D.
  Tard{\'o}s, ``Orb-slam3: An accurate open-source library for visual,
  visual--inertial, and multimap slam,'' \emph{IEEE Transactions on Robotics},
  vol.~37, no.~6, pp. 1874--1890, 2021.

\end{thebibliography}

\end{document}